# Context-based Image Segment Labeling (CBISL)

**Tobias Schlagenhauf**
Department of Mechanical Engineering
Karlsruhe Institute of Technology
Karlsruhe,76131

**Yefeng Xia**
Department of Mechanical Engineering
Karlsruhe Institute of Technology
Karlsruhe,76131

**Jürgen Fleischer**
Department of Mechanical Engineering
Karlsruhe Institute of Technology
Karlsruhe,76131

## Abstract

Working with images, one often faces problems with incomplete or unclear information. Image inpainting can be used to restore missing image regions but focuses, however, on low-level image features such as pixel intensity, pixel gradient orientation, and color. This paper aims to recover semantic image features (objects and positions) in images. Based on published gated PixelCNNs, we demonstrate a new approach referred to as quadro-directional PixelCNN to recover missing objects and return probable positions for objects based on the context. We call this approach context-based image segment labeling (CBISL). The results suggest that our four-directional model outperforms one-directional models (gated PixelCNN) and returns a human-comparable performance.

## 1     Introduction

Semantic segmentation paves the way for human and machine towards complete scene understanding and returns semantic image features. However, sometimes a segment image obtained from an image segmentation algorithm is not complete because of e.g. bad image conditions in reality, which cannot provide complete or reliable scene information. To solve the problem, a model is needed to label the segments whose labels are empty or current labels are doubtful in the segment image and where there is no access to any original image information. The model should depend on the other correctly labeled segment labels that provide contextual information in the image to make predictions on the unknown segments. We call this process context-based image segment labeling (CBISL).

With CBISL, many vision problems that occur in real-world situations are addressed, e.g. blind spots during vehicle driving. By tiling a pre-segmented image into N equal-sized blocks, we extract a block label matrix (BLM) from each of the segment blocks according to the principal class in the blocks to condense the relevant context information. In doing so, we gain semantic/context information about what is actually in the image in a human-interpretable sense. As the key to CBISL, we obtain a model which is able to predict existence together with the position of a class based on the existence and position of other classes in the image. In this way, we can perform e.g. semantic context completion if, for example, there are occluded regions for which we want to know the most probable class. Additionally, we can search for the most probable positions of different classes in an image based on the context information in the image. In doing so, we can draw probability maps for classes which do actually not exist in the image. An example could be the most probable position of a pedestrian in an image even if there is no pedestrian in the unknown region.

This is illustrated in the second row of Figure 1, which shows the probable position in the unknown region of pedestrians, vehicles, sidewalks, and traffic signs. The key is that these classes are predicted based on the context in the image.

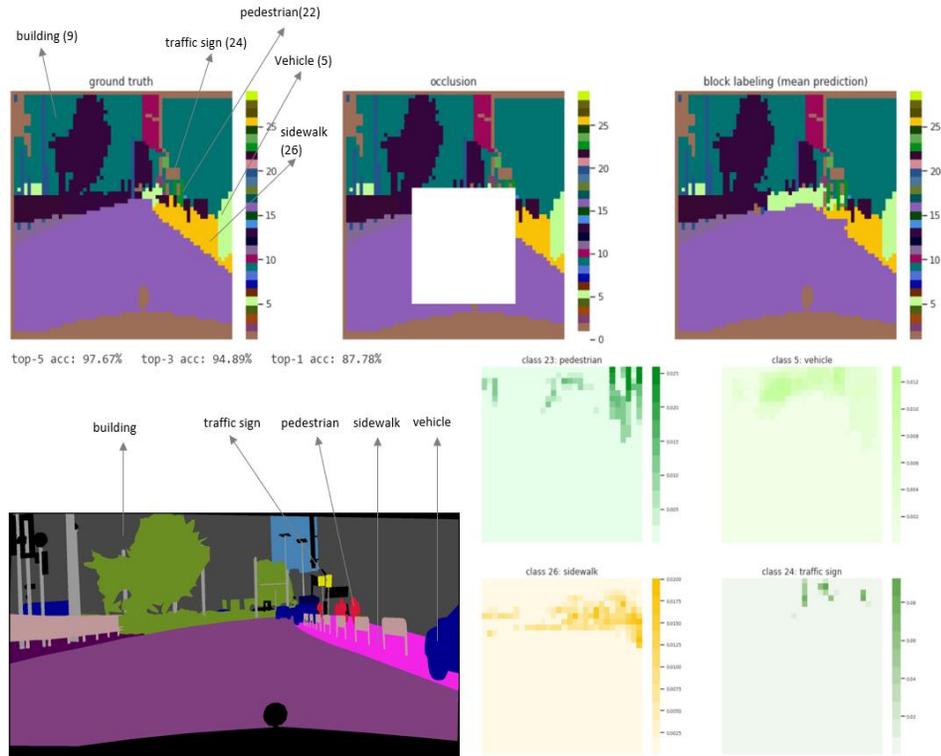

Figure 1 Result of CBISL on Cityscapes.

## 2 Background
### 2.1 Image Semantic Segmentation

Semantic segmentation is the task of clustering parts of images together which belong to the same object class (Szeliski 2010). Semantic image segmentation methods are usually divided into two categories: traditional and recent DNN methods. There are lots of progresses in backbone DNN, such as AlexNet architecture (Alex Krizhevsky, Ilya Sutskever, Geoffrey E. Hinton 2012), VGG-16 net (Simonyan, Karen, and Andrew Zisserman 2014), GoogLeNet (Christian Szegedy et al. 2015), ResNet (Deep residual network) (Kaiming He et al. 2016). Combined with the fully convolutional network (Jonathan Long, Evan Shelhamer, Trevor Darrell 2015), a label map can be obtained by the network by classifying every local region in the image. Conditional random field (CRF) is optionally applied to the output map for fine segmentation (Philipp Krähenbühl, Vladlen Koltun 2011). The deconvolution network is used to get rid of the problem that small objects are often ignored and classified as background when using FCNs. Atrous convolution (Chen u. a. 2018) contributes to detailed segmentation maps along object boundaries. Up until now, more and more methods have been emerging to make semantic image segmentation more accurate and/or faster.

### 2.2 Image Segment Labeling

For semantic segmentation, the segmentation map is pixel-level labeled output data. In order to represent a visual scene understanding for a segmented image, we divide the segmented image into m x n equally-sized blocks. A block gets labeled with the mode pixel class. Consequently, the segment image can be converted into block label matrices (BLM). In other words, BLM is an information-condensed version of the segmented image. However, when a segmented image is partially unknown, our approach may cope with the problem of incomplete information using the context information.

## 2.3 Autoregressive Models

Autoregressive models are generative models, which are known to be able to learn any kind of data distribution using unsupervised learning. Timely representatives of autoregressive models are PixelCNNs and PixelRNNs (Aäron van den Oord, Nal Kalchbrenner, Koray Kavukcuoglu 2016). Both approaches employ the basic building blocks of deep neural networks to formulate an approach to generate data, and also capture the full generality of pixel interdependencies. PixelCNNs/PixelRNNs model the joint distribution of pixels over an image as the following product of conditional distributions, where $x_i$ is a single pixel:

$$p_{(x)} = \prod_{i=1}^{n^2} p_{(x_i|x_1,\ldots,x_{i-1})}$$   Equation 2.1

In PixelCNNs, convolutional neural networks (CNN) can be used as sequence models with a fixed dependency range. PixelCNNs have a predefined sequence in raster scan order: row by row and pixel by pixel. Every pixel therefore conditions on all the pixels above and to the left of it. Therefore, two-dimensional autoregressive (AR-2D) models are one way to represent the image intensity of a given picture by a small number of parameters (Ojeda, Vallejos, Bustos 2010). Since the publications (Aaron van den Oord et al, 2016; Aäron van den Oord, Nal Kalchbrenner, Koray Kavukcuoglu 2016) presented PixelCNNs, they work well in the application of image inpainting (Ojeda, Vallejos, Bustos 2010; Britos et al. 2020; Uzan, Dershowitz, Wolf; Dupont, Emilien, and Suhas Suresha 2019). By using masked convolutions, PixelCNNs avoid seeing the future context which enables them to generalize based on the unmasked context. One drawback of PixelCNN is that each pixel only depends on the pixel before it while the pixels after it are ignored. Another drawback of PixelCNNs is the presence of a so called blind spot in the receptive field, where several pixels are left out (Aäron van den Oord, Nal Kalchbrenner, Koray Kavukcuoglu 2016). Although the pixels after it may be known also, they will not be taken into consideration by the model. This prevents it from obtaining the full context information of the image.

Compared with PixelRNNs, PixelRNNs generally have a better performance in spite of the fact that they are much faster to train because convolutions are inherently easier to parallelize.

PixelRNNs with spatial LSTM layers allow every layer in the network to access the entire neighborhood of previous pixels better than convolutional stacks in PixelCNNs (Aäron van den Oord, Nal Kalchbrenner, Koray Kavukcuoglu 2016). To improve this in PixelCNNs, the rectified linear units between the masked convolutions in the original PixelCNNs are replaced with the following gated activation unit:

$$y = \tanh(W_{k,f} * x) \odot \sigma(W_{k,g} * x),$$   Equation 2.2

where $\sigma$ is the sigmoid nonlinearity, k is the number of the layer, $\odot$ is the element-wise product, and * is the convolution operator.

Additionally, two stacks are used to overcome the blind spot, a horizontal stack (conditions only on the current row) and a vertical stack (conditions on all of the rows above) (Aaron van den Oordet al. 2016).

The here presented approach uses and extends the architecture of a gated PixelCNN by introducing a so-called quadro-directional PixelCNN which predicts unknown regions from all sides of the image to include maximum context information.

## 3 Similarity to Image Inpainting

Under a rough comparison, context-based image segment labeling is very close to image inpainting. Filling missing pixels of an image, often referred to as image inpainting or completion, is an important task in computer vision (Yu 2018). Image inpainting was applied on images in order to remove scratches and enhance damaged images. Now, it is used for removing artifact objects that can be added to the images by filling the target region with estimated values without using and returning important semantic context information about the included classes in the image. This is the twist in the here presented problem formulation. Due to the pre-segmentation (BLM), CBISL deals with additional human-interpretable information in advance. Both input and output in CBISL are represented as compressed-image contextual information while image inpainting models do not return context-relevant information in terms of existing classes as well as the position of these classes in an image. Additionally, CBISL can return the most probable positions of different classes in the target region of a segment image based on the context information which returns the probable position for all classes within the target region based on the class probability distribution over an image. Hence, image segment labeling is reformulated as a probability problem to return context-relevant information in terms of existing classes as well as the probable position of these classes in an image.

## 4    Approach

As introduced, autoregression is a time series model that uses observations from previous time steps as input to a regression equation to predict the value at the next time step.

To overcome the limitations of the PixelCNN stated above, we enlarge the model's utilization of sequence data, so that the model can predict the next pixel/block from its neighbor pixels/blocks of four different pixel sequences. Namely, there is not only one predefined sequence but four predefined sequences in total. We do not change the raster scan order in a neural network but change the data's sequence by rotating them. An image can be rotated three times, each time counterclockwise by 90 degrees. In doing so, we have four different pixel sequence images if the raster scan order is not changed always from top to bottom and left to right. We can obtain four kinds of train sets in such a way that all images on datasets are rotated three times similarly. Each train set can be used for training a gated PixelCNN with special pixel sequence. Quadro-directional PixelCNNs learn four different training pixel sequences that are from top left to bottom right, from top right to bottom left, from bottom left to top left, and from bottom right to top left. In so doing, we obtain four gated PixelCNNs which scan the pixel in different sequences. We refer to these gated PixelCNNs as gated_pixelCNN, gated_pixelCNN_90, gated_pixelCNN_180, and gated_pixelCNN_270 according to the image rotation degrees. We combine four gated PixelCNNs into one integrated-ensemble quadro-directional model. The four-directional model, like the normal gated PixelCNN, predicts the next pixel based on the known/predicted pixel value of previous vertical and current horizontal stacks, but the stacks lie at the top left, top right, bottom left, and bottom right of the next-to-predict pixel, thus the end prediction is a combined prediction of all four predictions. Figure 2 illustrates the prediction process for an unknown block in the BLM. In plot f, the four-directional model has not left out any known context information and the end prediction is actually based on all other previous models.

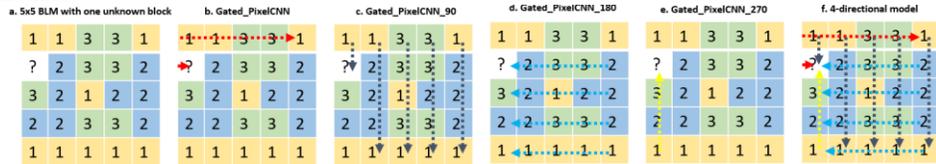

*Figure 2 All generation approaches in 5x5 BLM. a: An unknown/hidden block in the top left of the image. b: generation from gated_pixelcnn. c: generation from gated_pixelcnn_90. d: generation from gated_pixelcnn_180. e: generation from gated_pixelcnn_270. f: generation from four-directional model where all known blocks are applied.*

As a motivation that CBISL is useful for solving real-world vision problems, we apply our approach to two datasets. The first one is Cityscapes (Marius Cordts et al. 2016) made by Daimler AG R&D, TU Darmstadt, MPI Informatics, and TU Dresden. Cityscapes is a benchmark suite and large-scale dataset to train and test approaches for pixel-level and instance-level semantic labeling, which aims at visual understanding of complex urban street scenes. The second dataset is the COCO (common objects in context) dataset. COCO is a large-scale object detection, segmentation, and captioning dataset. A version of the COCO dataset is the COCO-Stuff dataset from 2017. The original COCO dataset is missing stuff annotations. To meet the demand of stuff annotations, the COCO-Stuff (Holger Caesar, Jasper Uijlings, Vittorio Ferrari 2018) is made by adding dense pixel-wise stuff annotations. This version is the version that is applied as our second dataset. In our experiments, we apply 3000 Cityscapes segment images and 118287 COCO-Stuff segment images as train sets, other 475 Cityscapes segment images and 5000 COCO-Stuff as test sets.

To condense the information in the original segment images, we created BLMs, size 64x64 for Cityscapes and 20x20 for COCO-Stuff. After training four times gated PixelCNN and integrating them into one four-directional model on both datasets, we test our model's performance on test sets, where our four-directional model is compared with each single gated PixelCNN. Table 1 records the negative likelihood loss (NLL) (Theis, Lucas, Aäron van den Oord, and Matthias Bethge 2015) of each sub-model and four-directional model on test sets. NLLs for the same dataset tend to be close, that on Cityscapes around 3.65 is only a little larger than the related publications. NLLs on COCO-Stuff are approximately 5.4 which is much larger, because of the larger number of classes and large variety in context compositions. Due to this high variability in context compositions, it is much more difficult to model the data density of BLMs on COCO-Stuff than on Cityscapes. With more image data available, the model should be able to also encode the semantic coherences in COCO-Stuff properly. This is the main limitation of this work. In general, the four-directional model outperforms each single gated PixelCNN, since the performance is based on the combined performance of its sub-models.

*Table 1 NLL of four sub-models and four-directional model on Cityscapes and COCO-Stuff.*

| NLL (bit/ dim) | Cityscapes | COCO-Stuff |
| --- | --- | --- |
| gated_pixelcnn | 3.6601 | 5.3846 |
| gated_pixelcnn_90 | 3.6544 | 5.3858 |
| gated_pixelcnn_180 | 3.6553 | 5.3817 |
| gated_pixelcnn_270 | 3.6558 | 5.3852 |
| 4-directional model | 3.6555 | 5.3837 |

## 5  Results & Discussion

In this chapter, our tests consider partially occluded COCO-Stuff and Cityscapes images. The occlusion area of images reaches from small to large at an arbitrary position. In consequence, BLM extracted from a segmented image is incomplete, where the labels of some blocks are unknown. On the one hand, four-directional models can fill out each unknown block with the most probable class. Additionally, our model can predict the probable occurrence of each class based on its spatial probability distribution in the target region. We take some tests on a GUI to compare our model against human predictions.

### 5.1  Image Segment Labeling on Cityscapes and COCO-Stuff

For the Cityscapes dataset, we set randomly 5x5, 20x20 and 40x40 unknown blocks on 100 test images and 2x2, 5x5 and 10x10 unknown blocks on 500 test images for COCO-Stuff.

*Table 2 Top-N Validation Accuracy (ACC) for CBISL on 100 Cityscapes segment images with five models.*

| Models on Cityscapes | Unknown blocks | Unknown area (%) | Top-5 acc (%) | Top-3 acc (%) | Top-1 acc (%) |
| --- | --- | --- | --- | --- | --- |
| Gated_pixelcnn | 5x5 | 0.61 | 96.96 | 94.64 | 82.32 |
| Gated_pixelcnn_90 | 5x5 | 0.61 | 93.2 | 88.24 | 74.04 |
| Gated_pixelcnn_180 | 5x5 | 0.61 | 93.56 | 89.48 | 74.80 |
| Gated_pixelcnn_270 | 5x5 | 0.61 | 91.16 | 86.48 | 72.32 |
| 4-directional model | 5x5 | 0.61 | 98.84 | 97.28 | 88.60 |
| Gated_pixelcnn | 20x20 | 9.8 | 86.32 | 80.56 | 65.82 |
| Gated_pixelcnn_90 | 20x20 | 9.8 | 73.66 | 67.56 | 53.26 |
| Gated_pixelcnn_180 | 20x20 | 9.8 | 83.26 | 75.16 | 55.98 |
| Gated_pixelcnn_270 | 20x20 | 9.8 | 70.38 | 62.58 | 43.92 |
| 4-directional model | 20x20 | 9.8 | 94.02 | 89.68 | 76.44 |
| Gated_pixelcnn | 40x40 | 39 | 79.46 | 72.04 | 54.30 |
| Gated_pixelcnn_90 | 40x40 | 39 | 49.38 | 42.38 | 27.90 |
| Gated_pixelcnn_180 | 40x40 | 39 | 78.54 | 68.32 | 48.00 |
| Gated_pixelcnn_270 | 40x40 | 39 | 48.56 | 41.78 | 27.84 |
| 4-directional model | 40x40 | 39 | 90.66 | 84.82 | 69.04 |

Table 3 Top-N ACC for CBISL on 500 COCO-Stuff segment images with five models.

| Models on COCO-Stuff | Unknown blocks | Unknown area (%) | Top-5 acc (%) | Top-3 acc (%) | Top-1 acc (%) |
|---|---|---|---|---|---|
| Gated_pixelcnn | 2x2 | 1 | 97.65 | 95.90 | 81.40 |
| Gated_pixelcnn_90 | 2x2 | 1 | 97.05 | 95.35 | 77.65 |
| Gated_pixelcnn_180 | 2x2 | 1 | 97.05 | 95.30 | 77.30 |
| Gated_pixelcnn_270 | 2x2 | 1 | 97.00 | 95.40 | 77.20 |
| 4-directional model | 2x2 | 1 | 99.50 | 98.50 | 88.35 |
| Gated_pixelcnn | 5x5 | 6.25 | 93.96 | 90.77 | 67.70 |
| Gated_pixelcnn_90 | 5x5 | 6.25 | 91.18 | 86.28 | 59.62 |
| Gated_pixelcnn_180 | 5x5 | 6.25 | 92.66 | 88.34 | 62.40 |
| Gated_pixelcnn_270 | 5x5 | 6.25 | 90.22 | 84.78 | 57.83 |
| 4-directional model | 5x5 | 6.25 | 98.33 | 95.68 | 80.62 |
| Gated_pixelcnn | 10x10 | 25 | 88.33 | 83.16 | 55.70 |
| Gated_pixelcnn_90 | 10x10 | 25 | 80.56 | 73.77 | 46.20 |
| Gated_pixelcnn_180 | 10x10 | 25 | 85.04 | 78.77 | 51.14 |
| Gated_pixelcnn_270 | 10x10 | 25 | 80.18 | 72.68 | 43.68 |
| 4-directional model | 10x10 | 25 | 95.29 | 90.70 | 71.85 |

From both Table 2 (Cityscpaes) and
Table 3 (COCO-Stuff), ACCs (Validation Accuracies) decrease with the increase in the unknown region's area. With the increase in the unknown region's area, top-5 and top-3 ACC stay relatively stable while the top-1 ACC decreases faster. The performance of a separated one-directional gated PixelCNN is always much worse than that of our four-directional model. The four-directional model's top-3 and top-5 accuracies are very stable and above 90% for all occluded region sizes in our test images. The top-1 ACC of the four-directional model is higher and decreases much more slowly than those of other models. The four-directional model outperforms the one-directional models on both datasets for all occluded region sizes. Figure 3 in the Appendix shows some examples of the tests. The predictions from the four-directional model at the last column in Figure 3 are close to ground truths in the second column.

## 5.2 Class Probabilities

Top-1 ACC as a criterion demonstrates our four-directional model's prediction within a small range of unknown blocks very precisely. Limitations arise when the unknown area increases to a larger region. However, deep autoregressive models return the probabilities for all classes in the unlabeled blocks during prediction. Based on tractable data density, there are logits for all classes per prediction which can be converted to probability maps. Unlike directly labeling a block with the highest probability among all classes, the approach here illustrates the probabilities for all classes in the occluded regions. We observe the spatial probability distribution of the classes which can be directly translated into the regions where the class would most probably exist. Pedestrian, vehicle, traffic light, and traffic sign are important components in street view understanding. Without seeing the real scene, we can only guess where they might appear based on limited experience and fuzzy memory. However, the four-directional model makes a reasonable guess based on tractable data density. The model returns the most probable position of the objects in the context of the target region. In the Appendix Figure 4 top, the probability map of pedestrians shows the probable position of a pedestrian in the street and the map of vehicles shows a car or a part of a car possibly in the left of the occluded region, where a dense point cloud lies. It is convincing that there may still be a driving vehicle following the car in front, but the color of this point cloud in the top right of the occluded region is not dark enough. Additionally, the probability map of pedestrians illustrates that the pedestrians are most likely to be walking on the sidewalk. Traffic sign and traffic light are probably in the middle of the road or on the sidewalk and away from the car hood. This idea works on COCO-Stuff as well. In the Appendix Figure 4 bottom, there are a man skiing in the snow and several trees connecting the sky and the snow ground. The prediction in the first row right is 84% top-1 ACC, nevertheless there is a segment block labeled as mountain between tree and snow blocks, which is false. Except the almost right predictions of clouds, sky, tree, snow, rock, and people segments, we can still see the probability maps of mountain and road. A probable position of the mountain is illustrated in the bottom left corner in the occluded region where the gap between clouds and snow ground is bridged. When people see trees and rocks, they may also think about mountains.

In summary, we can complete the image in a semantically correct way by filling in missing occluded classes. Additionally, the predictions have a human-interpretable semantic meaning such that we can now actually predict the most probable classes in this region, which is an information critical for scene understanding. Last but not least, we can search for the most probable positions of different classes in an image based on the context information in the image. We obtain all probable classes and positions explicitly illustrated by probability heat maps also for classes which do actually not exist in the image.

### 5.3 Human vs. Machine

To test our approach against the human intuition to predict classes based on their context, we created a GUI based on PyQt5, on which users can enter their predictions to fill in a BLM on Cityscapes or COCO-Stuff. With the GUI, we can compare the performance between the human and our model (machine). In our GUI experiments, we used ten test images each on Cityscapes and COCO-Stuff, whose occlusion regions individually consisted of 30x30 and 10x10 blocks. The comparison results are demonstrated in Table 4, where our four-directional model has a human-comparable performance on CBISL.

*Table 4 Performance comparison between user and model on GUI.*

| Cityscapes index (30x30 unknown) | user's ACC/ % | model's ACC/ % | COCO-Stuff index (10x10 unknown) | user's ACC/ % | model's ACC/ % |
|---|---|---|---|---|---|
| 1 | 71.00 | 62.00 | 1 | 98.00 | 98.00 |
| 2 | 62.00 | 63.00 | 2 | 98.00 | 85.00 |
| 3 | 73.00 | 72.00 | 3 | 99.00 | 99.00 |
| 4 | 97,00 | 94.00 | 4 | 100.00 | 100.00 |
| 5 | 85.00 | 89.00 | 5 | 71.00 | 84.00 |
| 6 | 74.00 | 68.00 | 6 | 95.00 | 96.00 |
| 7 | 72.00 | 59.00 | 7 | 96.00 | 97.00 |
| 8 | 88.00 | 89.00 | 8 | 67.00 | 67.00 |
| 9 | 96.00 | 94.00 | 9 | 91.00 | 93.00 |
| 10 | 62.00 | 60.00 | 10 | 83.00 | 84.00 |
| average | 78.00 | 75.00 | average | 89.00 | 90.00 |

Therefore, CBISL using the quadro-directional PixelCNN model can help with understanding contextual information in computer vision tasks by recovering scene information and providing useful information about the likely positions of classes.

### 6 Conclusion

In this paper we introduced our context based image segment labeling (CBISL) approach. We introduced an enhanced versions of Gated Pixel CNNs which, together with a reformulation of the image in painting problem, leads to new possibilities in the field of (context based) object detection and segmentation. We tested our approach on the well-known Coco-Stuff and Cityscapes dataset where we could predict the positions of classes in occluded regions based on the positions of other existing classes in the image. Additionally we tested our approach against the human ability to solve this problem. In this paper we firstly showed how to condense the class information in an image by building so called Block Label Matrices (BLM). In a second step we further developed PixelCNNs by introducing our quadro-directional PixelCNN approach. We then used the BLMs together with the quadro-directional PixelCNN model to learn the dependencies between the existence and the position of different classes in an image. In further works we will train the model with a larger database to receive more accurate results. We will further enhance the approach by enabling the model to extend the BLM of an image in each direction to gain class information about regions not visible in the image. In the future our approach could and should also be adopted to video material.

# Appendix

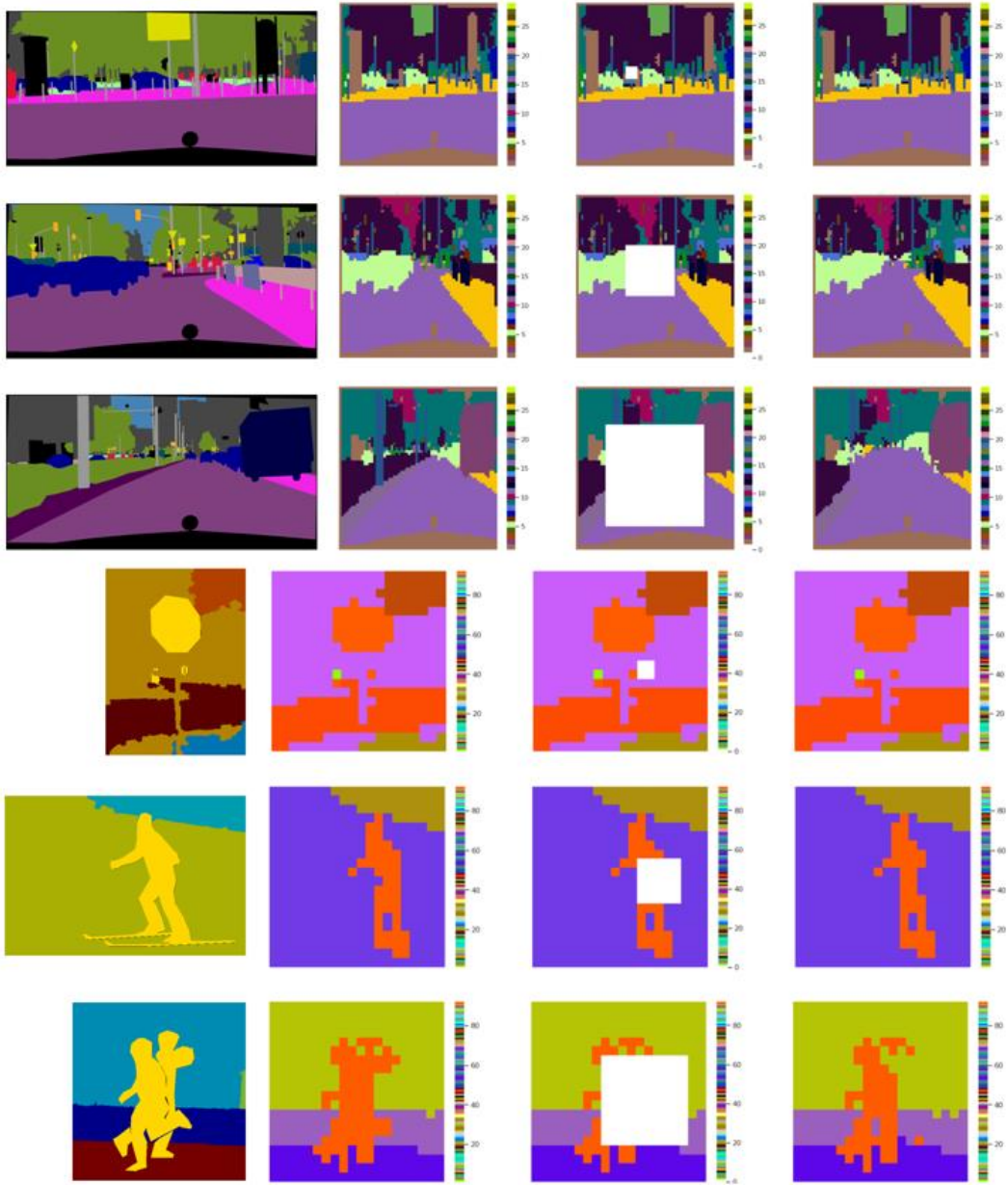

*Figure 3 Context-based image segment labeling on Cityscapes (top 3 rows) and COCO-Stuff (bottom 3 rows). First column: Ground truth. Second column: 64x64 BLM of Cityscapes plot (20x20 BLM of COCO-Stuff plot). Third column: Different amounts of unlabeled blocks in BLM. Fourth column: Our four-directional model output.*

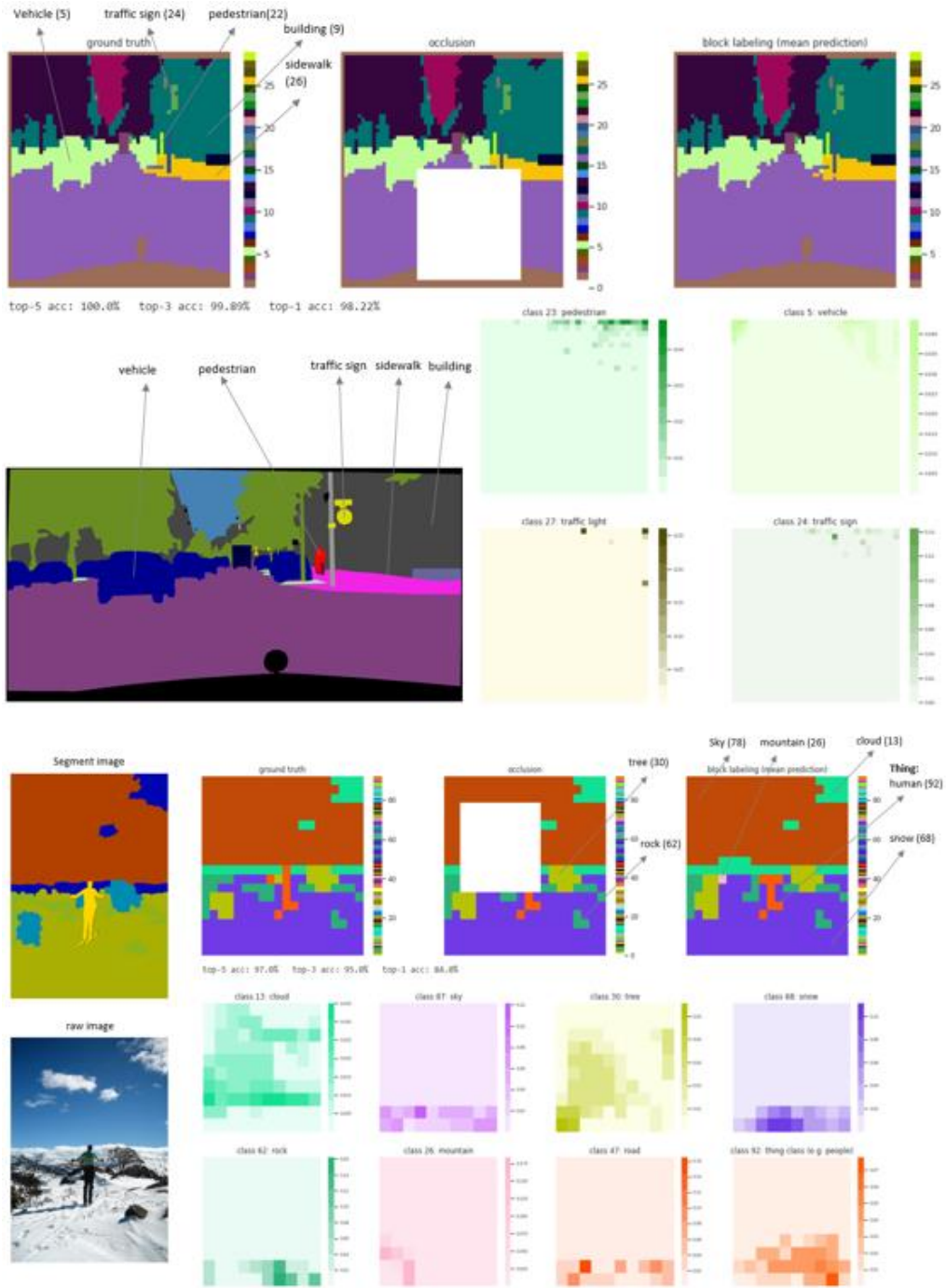

Figure 4 Predicting the probabilities for significant classes on Cityscapes/COCO-Stuff.